%% file: main-arxiv.tex
\documentclass[twocolumn]{article}
\PassOptionsToPackage{dvipsnames}{xcolor} 
\usepackage{preprint}
\SetBgContents{} 
\usepackage[utf8]{inputenc}
\usepackage[T1]{fontenc}
\usepackage[english]{babel}
\usepackage{csquotes}
\usepackage[final,
            tracking=true,
            kerning=true]{microtype}
\usepackage{xspace}

\usepackage{ifthen}
\newboolean{wide}
\setboolean{wide}{true}

\usepackage{amssymb}
\usepackage{amsmath}
\usepackage{amsthm}
\usepackage{mathtools}
\usepackage{bbm}

\usepackage{graphicx}
\usepackage{booktabs}
\usepackage{subcaption}
\usepackage{afterpage}
\usepackage{float}

\usepackage{enumitem}

\usepackage{nth}
\usepackage{siunitx}
\usepackage[dvipsnames]{xcolor}

\newcommand*{\titletext}{Adversarial attacks hidden in plain sight}

\usepackage[backend=biber,
            sorting=none,
            maxcitenames=2,
            maxbibnames=8]{biblatex}
\usepackage{varioref}
\usepackage[hypertexnames=false,
            colorlinks=true,
            linkcolor=MidnightBlue,
            urlcolor=Maroon,
            citecolor=ForestGreen,
            pdfauthor={Jan Philip Göpfert},
            pdftitle={\titletext}]{hyperref}
\usepackage[all]{hypcap}
\usepackage[noabbrev,
            capitalize,
            nameinlink]{cleveref}

\usepackage[backgroundcolor=white,
            linecolor=lightgray,
            bordercolor=white,
            textsize=tiny]{todonotes}

\input{macros.tex}

\title{\titletext}

\usepackage[auth-sc]{authblk}

\author[1]{Jan Philip Göpfert}
\author[1]{André Artelt}
\author[2]{Heiko Wersing}
\author[1]{Barbara Hammer}
\affil[1]{Bielefeld University, Germany}
\affil[2]{Honda Research Institute Europe GmbH, Offenbach, Germany}

\newcommand\blfootnote[1]{{%
  \let\thempfn\relax%
  \footnotetext[0]{#1}%
}}

\begin{document}
\twocolumn[
\begin{@twocolumnfalse}
\maketitle
\begin{abstract}
\input{abstract.tex}
\end{abstract}
\vspace{0.5cm}
\end{@twocolumnfalse}
]
\input{content.tex}
\input{acknowledgements.tex}
\AtNextBibliography{\raggedright\small}
\printbibliography
\input{appendix.tex}
\end{document}

%% file: macros.tex
\newcommand*{\eg}{e.\,g.\@\xspace}
\newcommand*{\ie}{i.\,e.\@\xspace}

\newcommand*{\N}{\mathbb{N}}
\DeclarePairedDelimiterX{\norm}[1]{\lVert}{\rVert}{#1}

\newcommand*{\gridelem}[2][]{%
\begin{subfigure}[t]{.3\linewidth}%
    \centering%
    \includegraphics[width=.9\linewidth]{media/#2.png}%
    \caption{#1}%
\end{subfigure}%
}%
\newcommand*{\gridcomplete}[1]{%
\gridelem[original input]{#1_original_mark}%
\gridelem[modified input, BIM]{#1_classic_attacked_mark}%
\gridelem[perturbation, BIM]{#1_classic_perturbation_mark}
\gridelem[original input (enlarged)]{#1_original_crop}%
\gridelem[modified input, BIM) (enlarged)]{#1_classic_attacked_crop}%
\gridelem[perturbation, BIM) (enlarged)]{#1_classic_perturbation_crop}
\gridelem[original input]{#1_original_mark}%
\gridelem[modified input, EbIM]{#1_entropy_attacked_mark}%
\gridelem[perturbation, EbIM]{#1_entropy_perturbation_mark}
\gridelem[original input (enlarged)]{#1_original_crop}%
\gridelem[modified input, EbIM (enlarged)]{#1_entropy_attacked_crop}%
\gridelem[perturbation, EbIM (enlarged)]{#1_entropy_perturbation_crop}%
}%

%% file: abstract.tex
Convolutional neural networks have been used to achieve a string of successes
during recent years, but their lack of interpretability remains a serious issue.
Adversarial examples are designed to deliberately fool neural networks into
making any desired incorrect classification, potentially with very high
certainty. Several defensive approaches increase robustness against adversarial
attacks, demanding attacks of greater magnitude, which lead to visible
artifacts. By considering human visual perception, we compose a technique that
allows to hide such adversarial attacks in regions of high complexity, such that
they are imperceptible even to an astute observer. We carry out a user study on
classifying adversarially modified images to validate the perceptual quality of
our approach and find significant evidence for its concealment with regards to
human visual perception.

%% file: content.tex
\section{Introduction}\label{struct:intro}
\begin{figure}[htp]%
    \centering%
    \gridcomplete{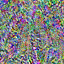}
    \caption{Two adversarial attacks carried out using the Basic Iterative
    Method (first two rows) and our Entropy-based Iterative Method (last two
    rows). The original image (a) (and (g)) is correctly classified as
    \emph{umbrella} but the modified images (b) and (h) are classified as
    \emph{slug} with a certainty greater than \SI{99}{\percent}. Note the
    visible artifacts caused by the perturbation (c), shown here with maximized
    contrast. The perturbation (i) does not lead to such artifacts. (d), (e),
    (f), (j), (k), and (l) are enlarged versions of the marked regions in (a),
    (b), (c), (g), (h), and (i), respectively.}%
    \label{fig:basic}%
\end{figure}
The use of convolutional neural networks has led to tremendous achievements
since \textcite{Krizhevsky2012ImageNet} presented AlexNet in~2012. Despite
efforts to understand the inner workings of such neural networks, they mostly
remain black boxes that are hard to interpret or explain. The issue was
exaggerated in~2013 when \textcite{Szegedy2013Intriguing} showed that
“adversarial examples” -- images perturbed in such a way that they fool a neural
network -- prove that neural networks do not simply generalize correctly the way
one might naïvely expect. Typically, such adversarial attacks change an input
only slightly, but in an adversarial manner, such that humans do not regard the
difference of the inputs relevant, but machines do. There are various types of
attacks, such as one pixel attacks, attacks that work in the physical world, and
attacks that produce inputs fooling several different neural networks without
explicit knowledge of those
networks~\cite{Su2017One,Kurakin2016Adversarial,Papernot2016Practical}.

Adversarial attacks are not strictly limited to convolutional neural networks.
Even the simplest binary classifier partitions the entire input space into
labeled regions, and where there are no training samples close by, the
respective label can only be nonsensical with regards to the training data, in
particular near decision boundaries. One explanation of the “problem” that
convolutional neural networks have is that they perform extraordinarily well in
high-dimensional settings, where the training data only covers a very thin
manifold, leaving a lot of “empty space” with ragged class regions. This creates
a lot of room for an attacker to modify an input sample and move it away from
the manifold on which the network can make meaningful predictions, into regions
with nonsensical labels. Due to this, even adversarial attacks that simply blur
an image, without any specific target, can be
successful~\cite{Chakraborty2018Adversarial}. There are further attempts at
explaining the origin of the phenomenon of adversarial examples, but so far, no
conclusive consensus has been
established~\cite{Goodfellow2014Explaining,Luo2015Foveation,Cisse2017Parseval,Ilyas2019Adversarial}.

A number of defenses against adversarial attacks have been put forward, such as
defensive distillation of trained networks~\cite{Papernot2016Distillation},
adversarial training~\cite{Madry2017Towards}, specific
regularization~\cite{Cisse2017Parseval}, and statistical
detection~\cite{Crecchi2019Detecting,Feinman2017Detecting,Grosse2017Statistical,Metzen2017Detecting}.
However, no defense succeeds in universally preventing adversarial
attacks~\cite{Carlini2017Towards,Athalye2018Obfuscated}, and it is possible that
the existence of such attacks is inherent in high-dimensional learning
problems~\cite{Chakraborty2018Adversarial}. Still, some of these defenses do
result in more robust networks, where an adversary needs to apply larger
modifications to inputs in order to successfully create adversarial examples,
which begs the question how robust a network can become and whether robustness
is a property that needs to be balanced with other desirable properties, such as
the ability to generalize well~\cite{Tsipras2019Robustness} or a reasonable
complexity of the network~\cite{Nakkir2019Adversarial}.

Strictly speaking, it is not entirely clear what defines an adversarial example
as opposed to an incorrectly classified sample. Adversarial attacks are devised
to change a given input minimally such that it is classified incorrectly -- in
the eyes of a human. While astonishing parallels between human visual
information processing and deep learning exist, as highlighted \eg by
\textcite{Yamins2016Using,Rajalingham2018Large}, they disagree when presented
with an adversarial example. Experimental evidence has indicated that specific
types of adversarial attacks can be constructed that also deteriorate the
decisions of humans, when they are allowed only limited time for their decision
making~\cite{Elsayed2018Adversarial}. Still, human vision relies on a number of
fundamentally different principles when compared to deep neural networks: while
machines process image information in parallel, humans actively explore scenes
via saccadic moves, displaying unrivaled abilities for structure perception and
grouping in visual scenes as formalized \eg in the form of the Gestalt
laws~\cite{Wersing2001Competitive,Ibbotson2011Visual,Lewicki2014Scene,Jaekel2016overview}.
As a consequence, some attacks are perceptible by humans, as displayed in
\cref{fig:basic}. Here, humans can detect a clear difference between the
original image and the modified one; in particular in very homogeneous regions,
attacks lead to structures and patterns which a human observer can recognize. We
propose a simple method to address this issue and answer the following
questions. How can we attack images using standard attack strategies, such that
a human observer does not recognize a clear difference between the modified
image and the original? How can we make use of the fundamentals of human visual
perception to “hide” attacks such that an observer does not notice the changes?

Several different strategies for performing adversarial attacks exist. For a
multiclass classifier, the attack's objective can be to have the classifier
predict \emph{any} label other than the correct one, in which case the attack is
referred to as \emph{untargeted}, or \emph{some specifically chosen} label, in
which case the attack is called \emph{targeted}. The former corresponds to
minimizing the likelihood of the original label being assigned; the latter to
maximizing that of the target label. Moreover, the classifier can be fooled into
classifying the modified input with extremely high confidence, depending on the
method employed. This, in particular, can however lead to visible artifacts in
the resulting images (see \cref{fig:basic}). After looking at a number of
examples, one can quickly learn to make out typical patterns that depend on the
classifying neural network. In this work, we propose a method for changing this
procedure such that this effect is avoided.

For this purpose, we extend known techniques for adversarial attacks. A
particularly simple and fast method for attacking convolutional neural networks
is the aptly named Fast Gradient Sign Method
(FGSM)~\cite{Goodfellow2014Explaining,Kurakin2016Adversarial}. This method, in
its original form, modifies an input image \(x\) along a linear approximation of
the objective of the network. It is fast but limited to untargeted attacks. An
extension of FGSM, referred to as the Basic Iterative Method
(BIM)~\cite{Kurakin2016Adversariala}, repeatedly adds small perturbations and
allows targeted attacks. \textcite{Moosavi-Dezfooli2016DeepFool} linearize the
classifier and compute smaller (with regards to the \(\ell_p\) norm)
perturbations that result in untargeted attacks. Using more computationally
demanding optimizations, \textcite{Carlini2017Towards} minimize the \(\ell_0\),
\(\ell_2\), or \(\ell_\infty\) norm of a perturbation to achieve targeted
attacks that are still harder to detect. \textcite{Su2017One} carry out attacks
that change only a single pixel, but these attacks are only possible for some
input images and target labels. Further methods exist that do not result in
obvious artifacts, \eg the Contrast Reduction Attack~\cite{Rauber2017Foolbox},
but these are again limited to untargeted attacks -- the input images are merely
corrupted such that the classification changes. None of the methods mentioned
here regard human perception directly, even though they all strive to find
imperceptibly small perturbations. \Textcite{Schoenherr2018Adversarial}
successfully do this within the domain of acoustics.

We rely on BIM as the method of choice for attacks based on images, because it
allows robust targeted attacks with results that are classified with arbitrarily
high certainty, even though it is easy to implement and efficient to execute.
Its drawbacks are the aforementioned visible artifacts. To remedy this issue, we
will take a step back and consider human perception directly as part of the
attack. In this work, we propose a straightforward, very effective modification
to BIM that ensures targeted attacks are visually imperceptible, based on the
observation that attacks do not need to be applied homogeneously across the
input image and that humans struggle to notice artifacts in image regions of
high local complexity. We hypothesize that such attacks, in particular, do not
change saccades as severely as generic attacks, and so humans perceive the
original image and the modified one as very similar -- we confirm this
hypothesis in \cref{sec:study} as part of a user study.
\section{Adversarial attacks}
\newcommand*{\zeroone}{\left[0, 1\right]}
\newcommand*{\mask}{\mathcal{E}}
\newcommand*{\ensum}{\kappa}
\newcommand*{\numbers}[1]{\overline{#1}}
Recall the objective of a targeted adversarial attack. Given a classifying
convolutional neural network \(f\), we want to modify an input \(x\), such that
the network assigns a different label \(f(x')\) to the modified input \(x'\)
than to the original \(x\), where the target label \(f(x')\) can be chosen at
will. At the same time, \(x'\) should be as similar to \(x\) as possible, \ie we
want the modification to be small. This results in the optimization problem:
\begin{equation}
    \min \norm{x' - x} \quad \text{such that} \quad f(x') = y \neq f(x),\label{eqn:adversarial}
\end{equation}
where \(y = f(x')\) is the target label of the attack. BIM finds such a small
perturbation \(x' - x\) by iteratively adapting the input according to the
update rule
\begin{equation}
    x \gets x - \epsilon \cdot \mathrm{sign}[\nabla_x J(x,y)]\label{eq:update}
\end{equation}
until \(f\) assigns the label \(y\) to the modified input with the desired
certainty, where the certainty is typically computed via the softmax over the
activations of all class-wise outputs. \(\mathrm{sign}[\nabla_x J(x,y)]\)
denotes the sign of the gradient of the objective function \(J(x,y)\), and is
computed efficiently via backpropagation; \(\epsilon\) is the step size. The
norm of the perturbation is not considered explicitly, but because in each
iteration the change is distributed evenly over all pixels/features in \(x\),
its \(\ell_{\infty}\)-norm is minimized.
\subsection{Localized attacks}
\begin{figure}[t]
    \centering
    \gridelem[modified input]{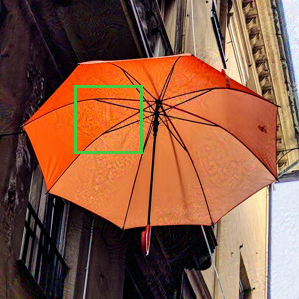}%
    \gridelem[modified input]{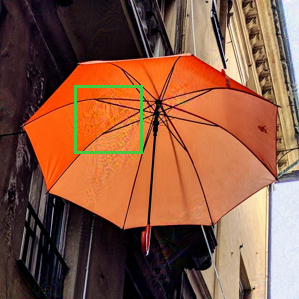}%
    \gridelem[modified input]{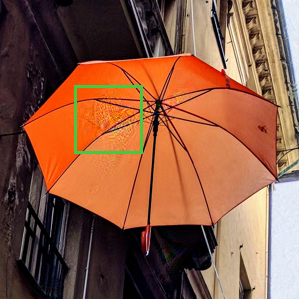}
    \gridelem[high strength]{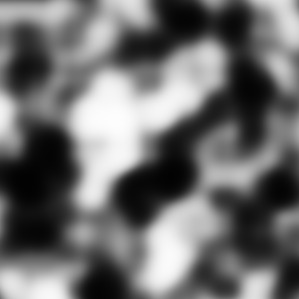}%
    \gridelem[medium strength]{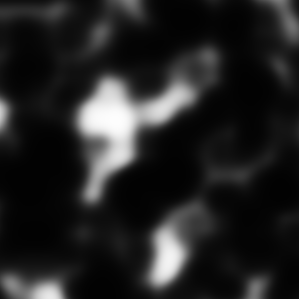}%
    \gridelem[low strength]{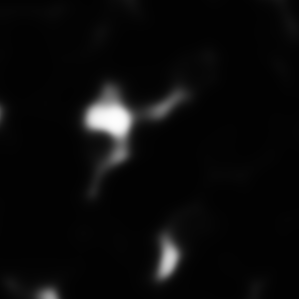}
    \gridelem[modified input (enlarged)]{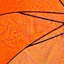}%
    \gridelem[modified input (enlarged)]{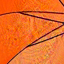}%
    \gridelem[modified input (enlarged)]{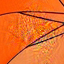}
    \caption{Localized attacks with different relative total strengths. The
    strength maps (d), (e), and (f), which are based on Perlin noise, scaled
    such that the relative total strength is \(0.43\), \(0.14\), and \(0.04\),
    are used to create the adversarial examples in (a), (b), and (c),
    respectively. In each case, the attacked image is classified as \emph{slug}
    with a certainty greater than \SI{99}{\percent}. The attacks took 14, 17,
    and 86~iterations. (g), (h), and (i) are enlarged versions of the marked
    regions in (a), (b), and (c).}
    \label{fig:perlin}
\end{figure}
The main technical observation, based on which we hide attacks, is the fact that
one can weigh and apply attacks locally in a precise sense: During prediction, a
convolutional neural network extracts features from an input image, condenses
the information contained therein, and conflates it, in order to obtain its best
guess for classification. Where exactly in an image a certain feature is located
is of minor consequence compared to how strongly it is
expressed~\cite{Sabour2017Dynamic,Brown2017Adversarial}. As a result, we find
that during BIM's update, it is not strictly necessary to apply the computed
perturbation evenly across the entire image. Instead, one may choose to leave
parts of the image unchanged, or perturb some pixels more or less than others,
\ie one may localize the attack. This can be directly incorporated into
\cref{eq:update} by setting an individual value for \(\epsilon\) for every
pixel.

For an input image \(x \in \zeroone^{w \times h \times c}\) of width \(w\) and
height \(h\) with \(c\) color channels, we formalize this by setting a strength
map \(\mask \in \zeroone^{w \times h}\) that holds an update magnitude for each
pixel. Such a strength map can be interpreted as a grayscale image where the
brightness of a pixel corresponds to how strongly the respective pixel in the
input image is modified. The adaptation rule~(\ref{eq:update}) of BIM is changed
to the update rule
\begin{equation}
    x_{ijk} \gets x_{ijk} - \epsilon \cdot \mask_{ijk} \cdot \mathrm{sign}[\nabla_x J(x,y)]\label{eq:updatechanged}
\end{equation}
for all pixel values \((i,j,k)\). In order to be able to express the overall
strength of an attack, for a given strength map \(\mask\) of size \(w\) by
\(h\), we call
\begin{equation}
    \ensum(\mask) = \frac{\sum_{i, j \in \numbers{w} \times \numbers{h}} \mask_{i, j}}{w \cdot h}
\end{equation}
the \emph{relative total strength} of \(\mask\), where for \(n \in \N\) we let
\(\numbers{n} = \{1, \dots, n\}\) denote the set of natural numbers from \(1\)
to \(n\). In the special case where \(\mask\) only contains either black or
white pixels, \(\ensum(\mask)\) is the ratio of white pixels, \ie the number of
attacked pixels over the total number of pixels in the attacked image.

As long as the scope of the attack, \ie \(\ensum(\mask)\), remains large enough,
adversarial attacks can still be carried out successfully -- if not as easily --
with more iterations required until the desired certainty is reached. This leads
to the attacked pixels being perturbed more, which in turn leads to even more
pronounced artifacts. Given a strength map \(\mask\), it can be modified to
increase or decrease \(\ensum(\mask)\) by adjusting its brightness or by
applying appropriate morphological operations. See \cref{fig:perlin} for a
demonstration that uses pseudo-random noise as a strength map.
\subsection{Entropy-based attacks}
The crucial component necessary for “hiding” adversarial attacks is choosing a
strength map \(\mask\) that appropriately considers human perceptual biases. The
strength map essentially determines which “norm” is chosen in
\cref{eqn:adversarial}. If it differs from a uniform weighting, the norm
considers different regions of the image differently. The choice of the norm is
critical when discussing the visibility of adversarial attacks. Methods that
explicitly minimize the \(\ell_p\) norm of the perturbation for some \(p\), only
“accidentally” lead to perturbations that are hard to detect visually, since the
\(\ell_p\) norm does not actually resemble \eg the human visual focus for the
specific image. We propose to instead make use of how humans perceive images and
to carefully choose those pixels where the resulting artifacts will not be
noticeable.

Instead of trying to hide our attack in the background or “where an observer
might not care to look”, we instead focus on those regions where there is high
local complexity. This choice is based on the rational that humans inspect
images in saccadic moves, and a focus mechanism guides how a human can process
highly complex natural scenes efficiently in a limited amount of time.
\emph{Visual interest} serves as a selection mechanism, singling out relevant
details and arriving at an optimized representation of the given stimuli
\cite{Carrasco2011Visual}. We rely on the assumption that adversarial attacks
remain hidden if they do not change this scheme. In particular, regions which do
not attract focus in the original image should not increase their level of
interest, while relevant parts can, as long as the adversarial attack is not
adding additional relevant details to the original image.

Due to its dependence on semantics, it is hard -- if not impossible -- to
agnostically compute the magnitude of \emph{interest} for specific regions of an
image. Hence, we rely on a simple information theoretic proxy, which can be
computed based on the visual information in a given image: the entropy in a
local region. This simplification relies on the observation that regions of
interest such as edges typically have a higher entropy than homogeneous regions
and the entropy serves as a measure for how much information is already
contained in a region -- that is, how much relative difference would be induced
by additional changes in the region.

Algorithmically, we compute the \emph{local entropy} at every pixel in the input
image as follows: After discarding color, we bin the gray values, \ie the
intensities, in the neighborhood of pixel \(i, j\) such that \(B_{i, j}\)
contains the respective occurrence ratios. The occurrence ratios can be
interpreted as estimates of the intensity probability in this neighborhood,
hence the local entropy \(S_{i,j}\) can be calculated as the Shannon entropy
\begin{equation}
    S_{i,j} = - \hspace{-.5em} \sum_{p \in B_{i, j}} p \log p .
\end{equation}
Through this, we obtain a measure of local complexity for every pixel in the
input image, and after adjusting the overall intensity, we use it as suggested
above to scale the perturbation pixel-wise during BIM's update. In other words,
we set 
\begin{equation}
    \mask = \phi(S) \label{eqn:adjustment}
\end{equation}
where \(\phi\) is a nonlinear mapping, which adjusts the brightness. The choice
of a strength map based on the local entropy of an image allows us to perform an
attack as straightforward as BIM, but localized, in such a way that it does not
produce visible artifacts, as we will see in the following experiments.

While we could attach our technique to any attack that relies on gradients, we
use BIM because of the aforementioned advantages including simplicity,
versatility, and robustness, but also because as the direct successor to FGSM we
consider it the most typical attack at present. As a method of performing
adversarial attacks, we refer to our method as the \emph{Entropy-based Iterative
Method (EbIM)}.
\section{A study of how humans perceive adversarial examples}\label{sec:study}
It is often claimed that adversarial attacks are imperceptible\footnote{We do
not want to single out any specific source for this claim, and it should not
necessarily be considered strictly false, because there is no commonly accepted
rigorous definition of what constitutes an adversarial example or an adversarial
attack, just as it remains unclear how to best measure adversarial robustness.
Whether an adversarial attack results in noticeable artifacts depends on a
multitude of factors, such as the attacked model, the underlying data
(distribution), the method of attack, and the target certainty.}. While this can
be the case, there are many settings in which it does not necessarily hold true
-- as can be seen in \cref{fig:basic}. When robust networks are considered and
an attack is expected to reliably and efficiently produce adversarial examples,
visible artifacts appear. This motivated us to consider human visual perception
directly and thereby our method. To confirm that there are in fact differences
in how adversarial examples produced by BIM and EbIM are perceived, we conducted
a user study with \num{35}~participants.
\subsection{Generation of adversarial examples}
To keep the course of the study manageable, so as not to bore our relatively
small number of participants, and still acquire statistically meaningful (\ie
with high statistical power) and comparable results, we randomly selected only
\num{20}~labels and \num{4}~samples per label from the validation set of the
\emph{ILSVRC 2012 classification challenge}~\cite{Russakovsky2015ImageNet},
which gave us a total of \num{80}~images. For each of these \num{80}~images we
generated a targeted high confidence adversarial example using BIM and another
one using EbIM -- resulting in a total of \num{240}~images. We set a fixed
target class and the target certainty to \num{0.99}. We attacked the pretrained
\emph{Inception~v3} model~\cite{Szegedy2016Rethinking} as provided by
\emph{keras}~\cite{chollet2015keras}. We set the parameters of BIM to \(\epsilon
= 1.0\), \(stepsize = 0.004\) and \(max\_iterations=1000\). For EbIM, we
binarized the entropy mask with a threshold of \(4.2\). We chose these
parameters such that the algorithms can reliably generate targeted high
certainty adversarial examples across all images, without requiring expensive
per-sample parameter searches.
\subsection{Study design}
For our study, we assembled the images in pairs according to \emph{three
different conditions}:
\begin{enumerate}[label=(\roman*),nosep]
\item The original image versus itself.
\item The original image versus the adversarial example generated by BIM.
\item The original image versus the adversarial example generated by EbIM.
\end{enumerate}
This resulted in \num{240} pairs of images that were to be evaluated during the
study.

All image pairs were shown to each participant in a random order -- we also
randomized the positioning (left and right) of the two images in each pair. For
each pair, the participant was asked to determine whether the two images were
identical or different. If the participant thought that the images were
identical they were to click on a button labeled “Identical” and otherwise on a
button labeled “Different” -- the ordering of the buttons was fixed for a given
participant but randomized when they began the study. To facilitate completion
of the study in a reasonable amount of time, each image pair was shown for
\num{5}~seconds only; the participant was, however, able to wait as long as they
wanted until clicking on a button, whereby they moved on to the next image pair.
\subsection{Hypotheses tests}
Our hypothesis was that it would be more difficult to perceive the changes in
the images generated by EbIM than by BIM. We therefore expect our participants
to click “Identical” more often when seeing an adversarial example generated by
EbIM than when seeing an adversarial generated by BIM.

\begin{figure}[H]
    \centering
    \ifthenelse{\boolean{wide}}%
        {\includegraphics[width=.8\linewidth]{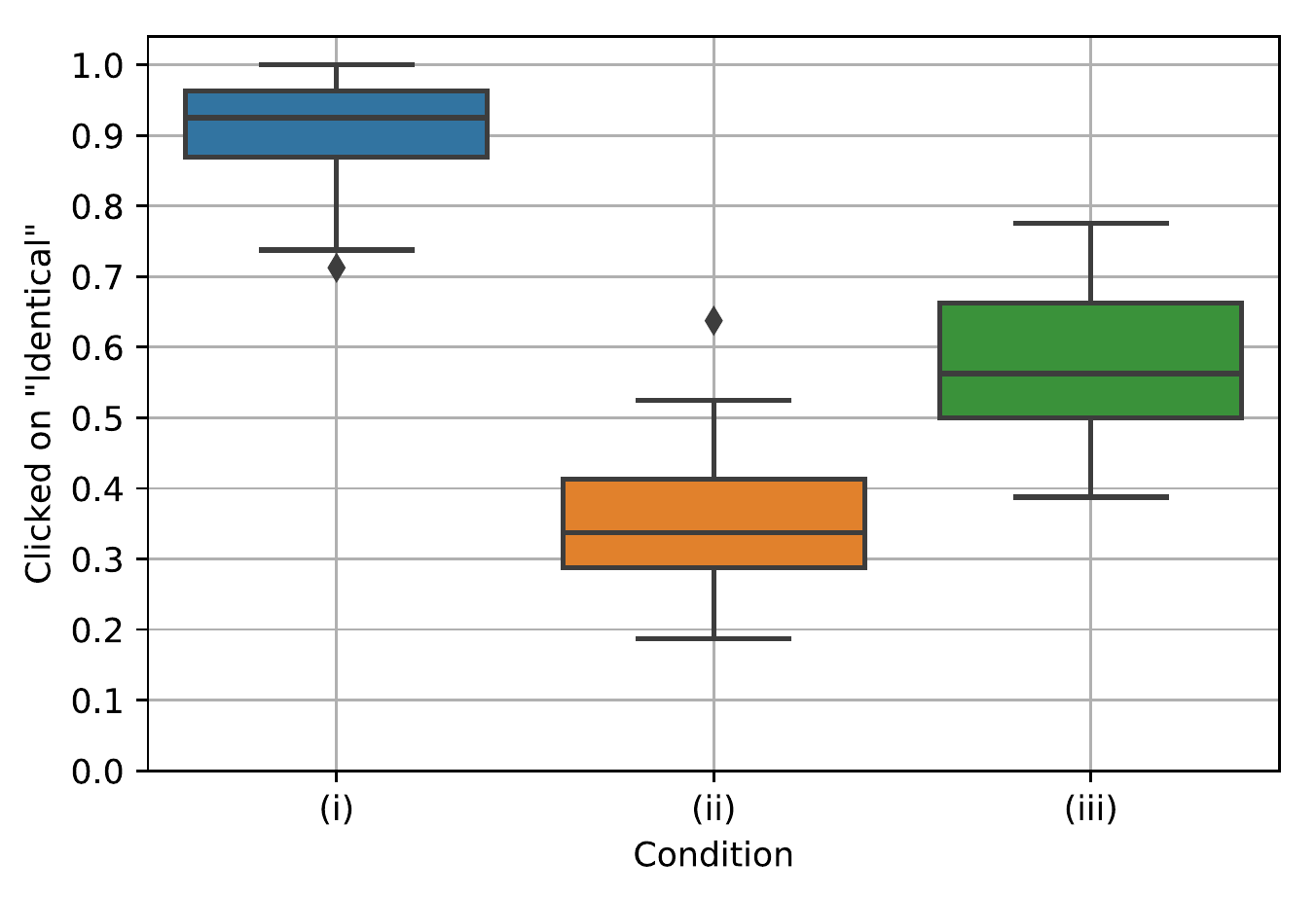}}%
        {\includegraphics[width=.5\linewidth]{media/study4.pdf}}
    \caption{Percentage of times users clicked on “Identical” when seeing two
    identical images (condition (i), blue box), a BIM adversarial (condition
    (ii), orange box), or an EbIM adversarial (condition (iii), green box).}
    \label{fig:study}
\end{figure}

As a test statistic, we compute \emph{for each participant} and \emph{for each
of the three conditions separately}, the percentage of time they clicked on
“Identical”. The values can be interpreted as a mean if we encode “Identical” as
\(1\) and “Different” as \(0\). Hereinafter we refer to these mean values as
\(\mu_{\text{BIM}}\) and \(\mu_{\text{EbIM}}\). For each of the three
conditions, we provide a boxplot of the test statistics in \cref{fig:study} --
the scores of EbIM are much higher than BIM, which indicates that it is in fact
much harder to perceive the modifications introduced by EbIM compared to BIM.
Furthermore, users almost always clicked on “Identical” when seeing two
identical images.

Finally, we can phrase our belief as a hypothesis test. We determine whether we
can reject the following five hypotheses:
\begin{enumerate}[label=(\arabic*),nosep]
    \item\label{test1} \(H_0 \colon \mu_{\text{BIM}} \geq \mu_{\text{EbIM}}\),
    \ie attacks using BIM are as hard or harder to perceive than EbIM.
    \item\label{test2} \(H_0 \colon \mu_{\text{BIM}} \geq 0.5\), \ie whether
    attacks using BIM are easier or harder to perceive than a random prediction
    \item\label{test3} \(H_0 \colon \mu_{\text{EbIM}} \leq 0.5\), \ie whether
    attacks using EbIM are easier or harder to perceive than a random prediction
    \item\label{test4} \(H_0 \colon \mu_{\text{BIM}} \geq \mu_{\text{NONE}}\),
    \ie whether attacks using BIM are as easy or easier to perceive than
    identical images.
    \item\label{test5} \(H_0 \colon \mu_{\text{EbIM}} \geq \mu_{\text{NONE}}\),
    \ie whether attacks using EbIM are as easy or easier to perceive than
    identical images.
\end{enumerate}

We use a \emph{one-tailed t-test} and the (non-parametric) \emph{Wilcoxon signed
rank test} with a significance level \(\alpha = 0.05\) in both tests. The
cases~\ref{test1},\ref{test4} and~\ref{test5} are tested as a \emph{paired test}
and the other two cases~\ref{test2} and~\ref{test3} as \emph{one sample tests}.

Because the t-test assumes that the mean difference is normally distributed, we
test for normality\footnote{Because we have \num{35}~participants, we assume
that normality approximately holds because of the central limit theorem.} by
using the \emph{Shapiro-Wilk normality test}. The Shapiro-Wilk normality test
computes a p-value of \num{0.425}, therefore we assume that the mean difference
follows a normal distribution. The resulting p-values are listed in
\vref{table:study} -- we can reject all null hypotheses with very low
p-values.
\begin{table*}
    \caption{p-values of each hypothesis (columns) under each test (rows). We
    reject all null hypotheses.}\label{table:study}
    \centering
    \sisetup{
        table-number-alignment=center,
        table-format=1.2e-1,
        round-mode=places,
        round-precision=2
    }
    \begin{tabular}{lSSSSS}\toprule
    Test     & {Hyp.~\ref{test1}} & {Hyp.~\ref{test2}} & {Hyp.~\ref{test3}} & {Hyp.~\ref{test4}} & {Hyp.~\ref{test5}} \\\midrule
    t-test   &            2.2e-16 &          1.025e-10 &          2.129e-05 &            2.2e-16 &            2.2e-16 \\ 
    Wilcoxon &          1.281e-07 &          9.101e-07 &          6.748e-05 &          1.283e-07 &          1.282e-07 \\\bottomrule
    \end{tabular}
\end{table*}

In order to compute the power of the t-test, we compute the effect size by
computing \emph{Cohen's d}. We find that \(d \approx 2.29\) which is considered
a huge effect size~\cite{Sawilowsky2009}. The power of the one-tailed t-test is
then approximately \(1\).

We have empirically shown that adversarial examples produced by EbIM are
significantly harder to perceive than adversarial examples generated by BIM.
Furthermore, adversarial examples produced by EbIM are not perceived as
differing from their respective originals.
\section{Discussion}
Adversarial attacks will remain a potential security risk on the one hand and an
intriguing phenomenon that leads to insight into neural networks on the other.
Their nature is difficult to pinpoint and it is hard to predict whether they
constitute a problem that will be solved. To further the understanding of
adversarial attacks and robustness against them, we have demonstrated two key
points:
\begin{itemize}
    \item Adversarial attacks against convolutional neural networks can be
    carried out successfully even when they are localized.
    \item By reasoning about human visual perception and carefully choosing
    areas of high complexity for an attack, we can ensure that the adversarial
    perturbation is barely perceptible, even to an astute observer who has
    learned to recognize typical patterns found in adversarial examples.
\end{itemize}

This has allowed us to develop the Entropy-based Iterative Method (EbIM), which
performs adversarial attacks against convolutional neural networks that are hard
to detect visually even when their magnitude is considerable with regards to an
\(\ell_p\)-norm. It remains to be seen how current adversarial defenses perform
when confronted with entropy-based attacks, and whether robust networks learn
special kinds of features when trained adversarially using EbIM.

Through our user study we have made clear that not all adversarial attacks are
imperceptible. We hope that this is only the start of considering human
perception explicitly during the investigation of deep neural networks in
general and adversarial attacks against them specifically. Ideally, this would
lead to a concise definition of what constitutes an adversarial example.

%% file: acknowledgements.tex
\section*{Acknowledgements}
We gratefully acknowledge support by Honda Research Institute Europe, funding
from the VW-Foundation for the project \emph{IMPACT} funded in the frame of the
funding line \emph{AI and its Implications for Future Society}, and we thank
Thomas Cederborg, Christina Göpfert, Benjamin Paaßen, and Ricarda Wullenkord for
helpful discussions.

%% file: appendix.tex
\section*{Appendix}
\begin{figure}
    \centering
    \gridelem{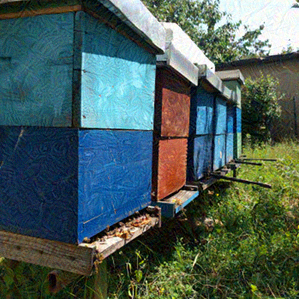}%
    \gridelem{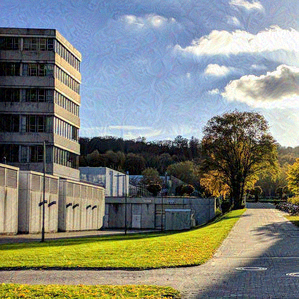}%
    \gridelem{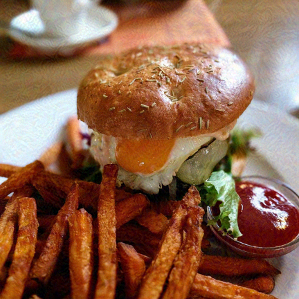}
    \gridelem{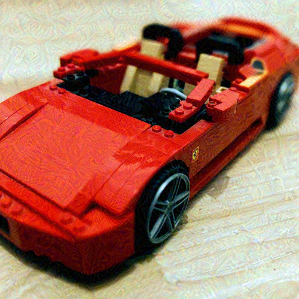}%
    \gridelem{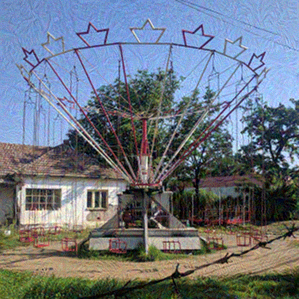}%
    \gridelem{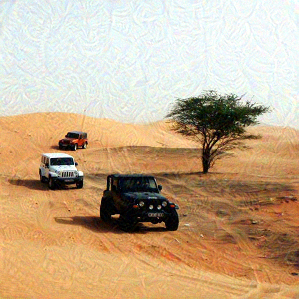}
    \gridelem{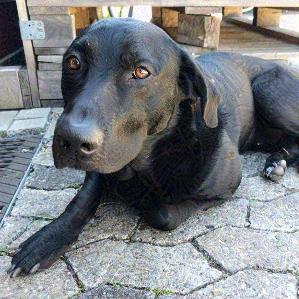}%
    \gridelem{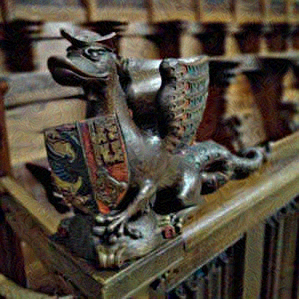}%
    \gridelem{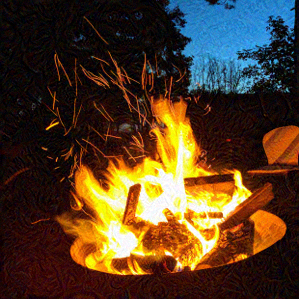}
    \gridelem{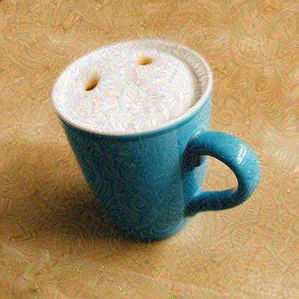}%
    \gridelem{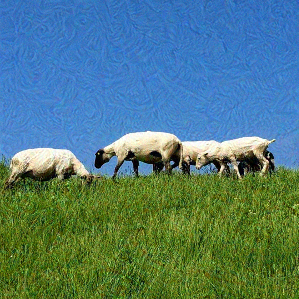}%
    \gridelem{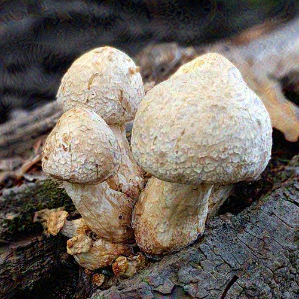}
    \caption{Twelve adversarial example resulting from attacks against Inception~v3 using the Basic Iterative Method.}
    \label{fig:inception-fgsm}
\end{figure}

\begin{figure}
    \centering
    \gridelem{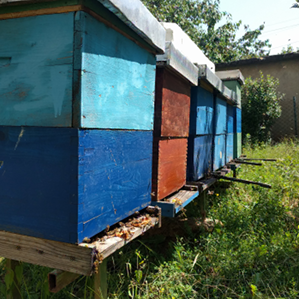}%
    \gridelem{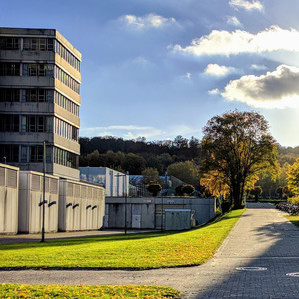}%
    \gridelem{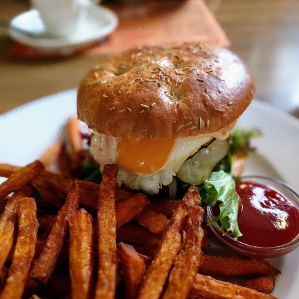}
    \gridelem{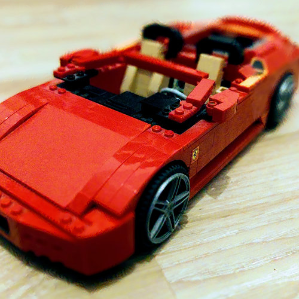}%
    \gridelem{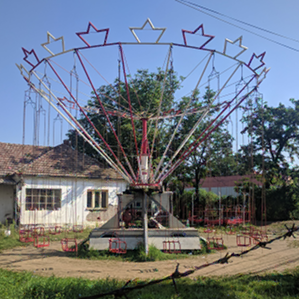}%
    \gridelem{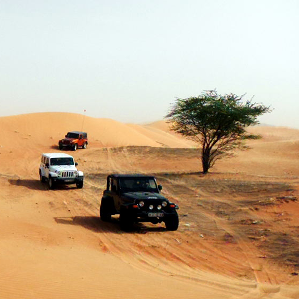}
    \gridelem{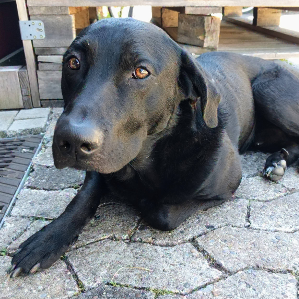}%
    \gridelem{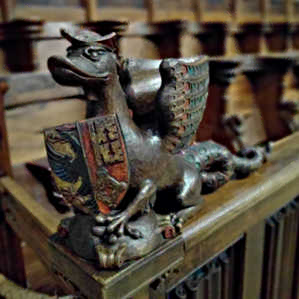}%
    \gridelem{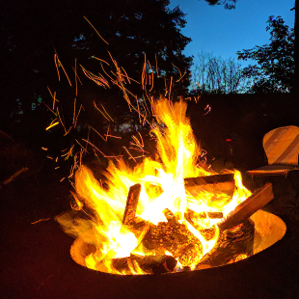}
    \gridelem{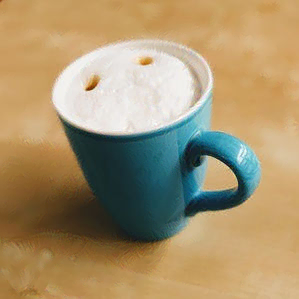}%
    \gridelem{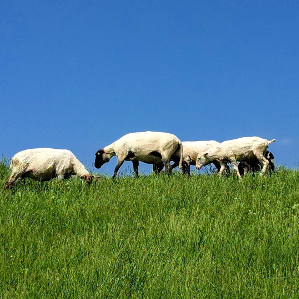}%
    \gridelem{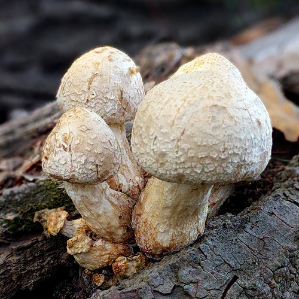}
    \caption{Twelve adversarial example resulting from attacks against Inception~v3 using our Entropy-based Iterative Method.}
    \label{fig:inception-entropy}
\end{figure}

\begin{figure}
    \centering
    \gridelem{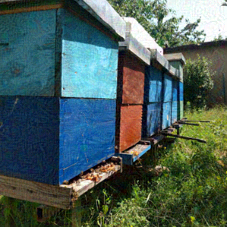}%
    \gridelem{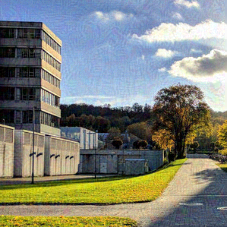}%
    \gridelem{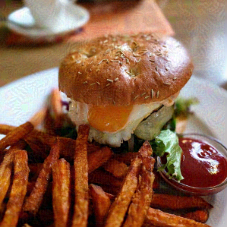}
    \gridelem{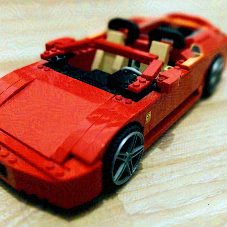}%
    \gridelem{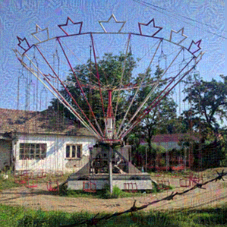}%
    \gridelem{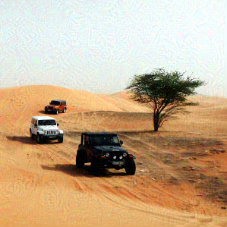}
    \gridelem{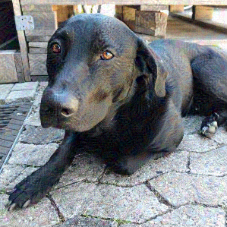}%
    \gridelem{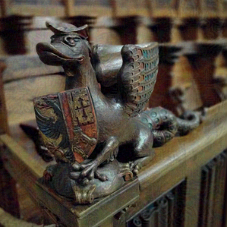}%
    \gridelem{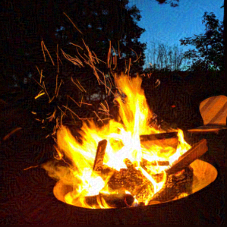}
    \gridelem{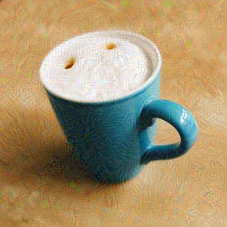}%
    \gridelem{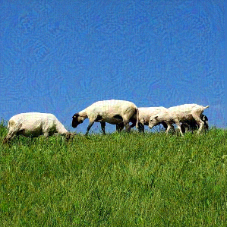}%
    \gridelem{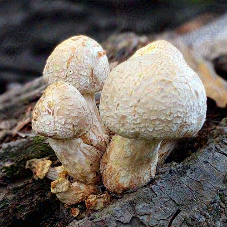}
    \caption{Twelve adversarial example resulting from attacks against AlexNet using the Basic Iterative Method.}
    \label{fig:alexnet-fgsm}
\end{figure}

\begin{figure}
    \centering
    \gridelem{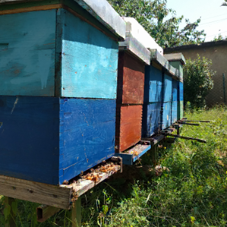}%
    \gridelem{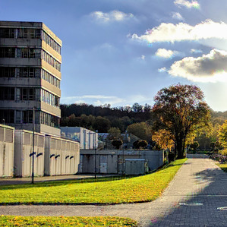}%
    \gridelem{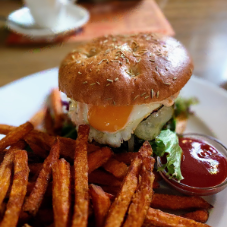}
    \gridelem{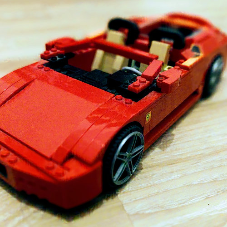}%
    \gridelem{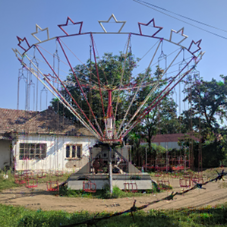}%
    \gridelem{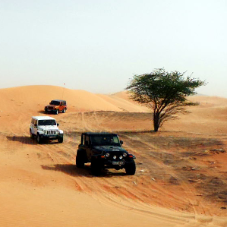}
    \gridelem{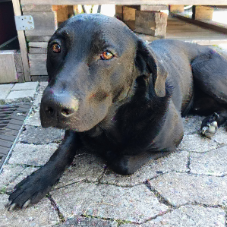}%
    \gridelem{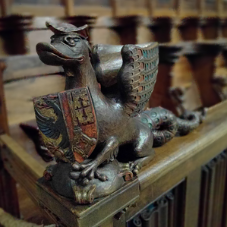}%
    \gridelem{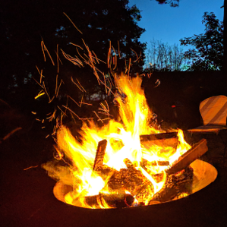}
    \gridelem{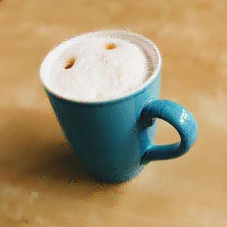}%
    \gridelem{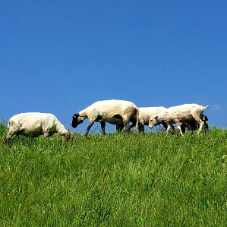}%
    \gridelem{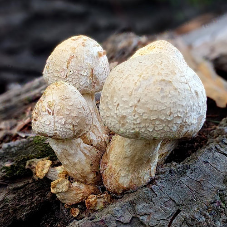}
    \caption{Twelve adversarial example resulting from attacks against AlexNet using our Entropy-based Iterative Method.}
    \label{fig:alexnet-entropy}
\end{figure}

For illustrative purposes, we include a number of adversarial examples that
result from attacks using the Basic Iterative Method (BIM) and our Entropy-based
Iterative Method (EbIM), against Inception~v3 and AlexNet, in
\vref{fig:inception-fgsm,fig:inception-entropy,fig:alexnet-fgsm,fig:alexnet-entropy}.
The images are best viewed digitally on a screen, with the ability to zoom in.
It should become clear how BIM leads to perceivable artifacts in low-entropy
regions, which is avoided when EbIM is employed. At the same time, certain
(comparably hard to see) artifacts result from EbIM, too, particularly around
pronounced edges. Note that this effect might be mitigated by smarter choices
with regards to the morphological operations applied to the pixel-wise entropy
before it is used as a strength map.
\clearpage
